# Beyond Theory of Mind in Robotics

Malte F. Jung

## Summary

"Robots need not read minds: social meaning emerges through interaction, calling for a shift beyond the Theory of Mind paradigm."

## Abstract

Theory of Mind (ToM) — the capacity to explain and predict behavior by inferring hidden mental states — has become the dominant paradigm for social interaction in robotics. Yet ToM rests on three assumptions that poorly capture how most social interaction actually unfolds: that meaning travels inside-out from hidden states to observable behavior; that understanding requires detached inference rather than participation; and that the meaning of behavior is fixed and available to a passive observer. Drawing on ethnomethodology, conversation analysis, and participatory sense-making, I argue that social meaning is not decoded from behavior but produced through moment-to-moment coordination between agents. This interactional foundation has direct implications for robot design: shifting from internal state modeling toward policies for sustaining coordination, from observer-based inference toward active participation, and from fixed behavioral meaning toward meaning potential stabilized through response.

# Introduction

You are walking down a hallway when someone approaches from the opposite direction. As you pass, the person raises their hand and moves it side to side. You recognize the motion as a wave, infer that the person intends to greet you, and respond by waving back.

This is how Theory of Mind (ToM) would have us believe social interaction works: we observe behavior, infer an underlying mental state or intention, and then select an appropriate response. Social understanding, on this view, is a process of reverse-engineering what is happening inside another person's head and acting accordingly. This logic persists even when ToM is not named directly. In many areas of robotics and human-robot interaction, including joint action and collaboration, social navigation, or emotion expression, systems are built on the premise that to understand behavior we need to infer the internal states that caused them.

In this paper, I argue that this paradigm is limited in the situations it applies to. It rests on assumptions about social interaction that do not adequately capture how interaction actually takes place. Drawing from research in ethnomethodology and conversation analysis as well as participatory sense-making, I argue for an interactional foundation for social interaction in robotics: a foundation that foregrounds moment-to-moment sense making in interaction.

Despite what the provocative title may suggest, I do not intend to dismiss theory of mind and its value for robotics. My aim is different here: While Theory of Mind has led to groundbreaking advances in robotics, its underlying assumptions will lead to a plateau in terms of the social situations robots can handle. A growing body of current robotics research has already departed from traditional ToM approaches towards data driven approaches that do not rely on internal state inference. What is missing, is a conceptual foundation to ground this burgeoning work in that will help it to develop further. An outline for such a conceptual foundation is the aim of this paper.

## Theory of Mind in Robotics

Theory of Mind (ToM) refers to the *capacity* to explain, predict, and justify the behavior of others by ascribing mental states such as goals, intentions, beliefs, or desires to them. Penn and Povinelli (2008, p. 394) define ToM broadly as: "any cognitive system, whether theory-like or not, that predicts or explains the behavior of another agent by postulating that unobservable inner states particular to the cognitive perspective of that agent causally modulate that agent's behavior." ToM is often used alongside related concepts such as intentional stance—the cognitive *strategy* of treating an entity as having goals and intentions (Dennett, 1989)—and mentalizing or mind reading, which refer to the actual *process* of inferring the internal states of others (Frith & Frith, 1999).

Theory of Mind was first studied systematically in primates (Premack & Woodruff, 1978) and subsequently in children. Wimmer and Perner's influential false-belief task (1983) provided an empirical milestone in ToM research. In this paradigm, children observe one character place an object in a location and leave the scene; a second character then moves the object. The child is asked where the first character will look for the object upon returning. This experimental design is telling: it establishes understanding as a matter of inferring hidden mental content from observed behavior, with the child as an uninvolved observer removed from actual interaction. The false-be-lief task became the canonical measure of ToM, and importantly, it embedded a particular epistemology into the very foundations of the field: understanding means reverse-engineering what is happening inside another agent's mind based on behavioral observation alone. Having Theory of Mind is widely regarded as a prerequisite for social intelligence in humans (Byom & Mutlu, 2013), and roboticist interested in developing robots capable of social interaction soon took notice.

Scassellati's highly influential 2002 paper on a "Theory of Mind for a Humanoid Robot" took inspiration from the work of Baron-Cohen (1997) and Frith and Frith (1999) and is generally regarded as the first systematic attempt to import developmental theory-of-mind models into robotics. Drawing on Scassellati's foundational work, Theory of Mind like capabilities have been used in robotics and HRI to advance approaches for dialogue understanding (e.g. Lemaignan et al., 2012), human-robot collaboration and teamwork (e.g. Hiatt, Harrison, & Trafton, 2011), social gaze and emotion (e.g. Breazeal et el., 2005). Theory of Mind soon became widely regarded as the crucial building block for building robots with social capabilities. Thus it is no surprise that Science Robotics listed Theory of Mind for robots as one of the grand challenges of robotics (Yang et al., 2018).

## Theory of Mind as a Broader Paradigm for Behavior Interpretation in Robotics.

The core premise of Theory of Mind—that we must reason about others' internal states (e.g., beliefs, intentions, emotions) to predict and explain their behavior in interaction and collaboration—extends beyond research that explicitly cites it. It rests on assump-tions that form a broader paradigm which Leudar and Costall (2009) called ToMism. Three core assumptions include inside-out, mentalizing, and sufficiency.

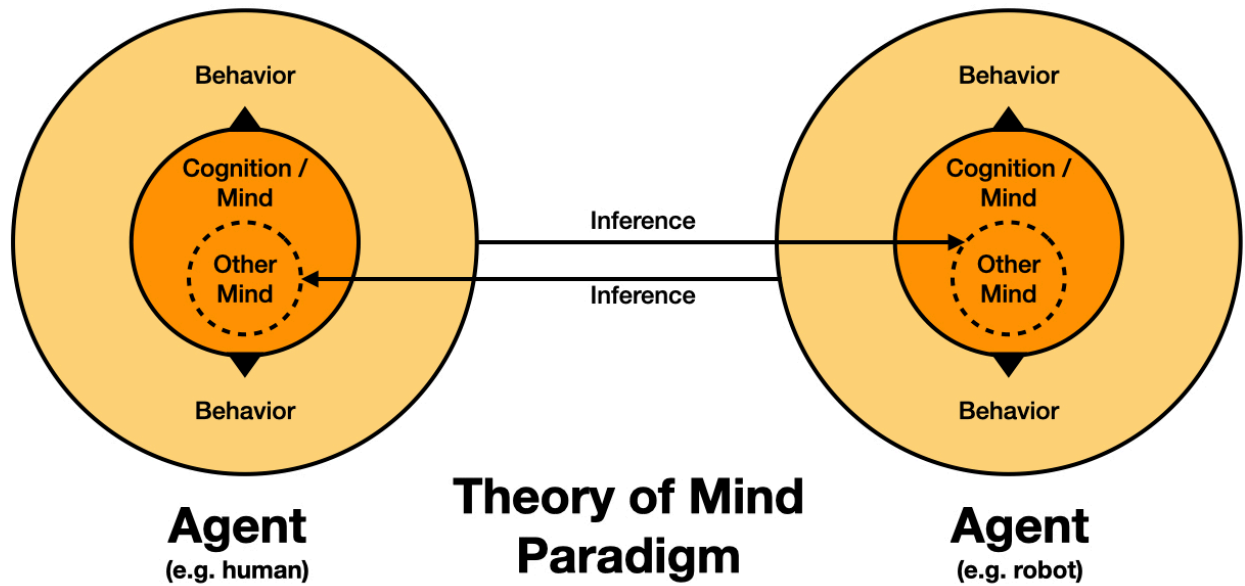

*Figure 1. A conceptual visualization of two agents interacting with each other consistently with the Theory of Mind paradigm. In this paradigm, meaning travels inside-out: hidden internal states and processes (Cognition / Mind) cause behavior, and understanding requires inferring the other's mind from observable behavior. Note. I am using a dyadic interaction example here only for illustration. The same paradigm holds for multi-party interactions.*

**Inside-Out:** Grounded in cartesian logic, the inside-out assumption holds that meaning travels inside-out - from hidden internal states and processes to the outward behavior that we can observe and interpret. Making sense of others' behavior then means to understand the hidden internal states that caused it. As Leudar and Costall (2009, p.5) put it: "In making sense of one another, we need to bridge a gulf between what we can 'directly' experience about other people, and what is going on 'in' their minds: 'our sensory experience of other people tells us about their movements in space but does not tell directly about their mental states' (Leudar and Costall, 2009, p. 5). This logic follows classic information theoretic views where information (internal state) is encoded into a message (behavior) by a sender and has to be decoded to be understood by a receiver (Shannon, 1948; Cha et al., 2018).  This implies that for a robot to understand, predict, and generate behavior it needs to have a model of the internal states of the agent it is interacting with. Reversely, this implies that for robots to be understandable they need to reveal the internal and hidden states that caused their behavior - making behavior legible, explainable, and specifically expressive for people to understand.

**Mentalizing:** The mentalizing assumption holds that making sense of others' behavior can only be achieved through some form of inference - as opposed to interaction, for example. From this view, understanding one another requires us to solve the "other minds problem" of knowing "what is going on in another person's mind" (Malle, 2007; p. 1) through "inference, theorizing, simulation or some other kind of 'detour'" (Leudar and Costall, 2009, p.5). This assumption presumes that sense making does not require interaction and instead treats it as a problem of decoding. It posits an asymmetrical relationship between understander and understood and places the responsibility of un-derstanding solely on the observer. For robotics, this implies that to equip robots with the capacity to engage in coordinated interactions with people means to equip them with the ability to model the internal states of others and make inferences based on observable behavior.

**Sufficiency:** The sufficiency assumption holds that the meaning of behavior is already contained in the behavior and its surrounding context, waiting to be decoded. This means that everything necessary to understand a social situation is available to a passive observer. This assumption is reflected in the classic false belief task (Wimmer & Perner, 1983), where the child is expected to interpret an unfolding situation as an uninvolved observer. For robotics, this implies that everything a robot needs to derive meaning from is available in the behavior and its context. Understanding behavior then becomes a data availability problem. Reversely this means that for a robot to be understandable or to make sense to others, it needs to encode all relevant information in the behavior and make it inherently legible or explain able. The robot's job is done once the behavior is generated and "expressed."

These assumptions underlie work in robotics and HRI that extend far beyond research that explicitly names ToM. They permeate several core problem domains in robotics. For example, most research on joint action, human-robot teaming and collaboration is consistent with inside-out and mentalizing assumptions which imply that successful collaboration rests on abilities to infer or model internal states, goals or intentions (e.g. Lasota, Fong, & Shah, 2017, Hoffman, Bhattacharjee, & Nikolaidis, 2024). Research on social, affective and expressive robotics is largely consistent with inside-out and sufficiency assumptions as it treats specific behaviors as inherently expressive or communicative, and views the primary function of expressive behavior as the communication of internal states (e.g. Dragan, Lee, & Srinivasa, 2013, Jung, 2017, Cha et al., 2018,

Admoni, & Scassellati, 2017). Finally, work on legibility, predictability, and explainability is largely consistent with inside-out and mentalizing assumptions which imply that alignment is a process of adapting internal and hidden states and that explainability is an expression or inference problem rather than viewing it as an interactional achievement (Dragan, Lee, & Srinivasa, 2013, Thellman & Ziemke, 2021, Mavrogiannis et al., 2023). Importantly, these assumptions persist even in contemporary data-driven systems that do not explicitly infer mental states, for example, by treating behavior interpretation as an inference problem performed from the perspective of an observer rather than as a joint achievement performed through interaction.

As a whole, research grounded in ToM assumptions has led to major advances in robotics that have enabled machines to better understand, predict and respond to human behavior. However, it has contributed to a dominant approach in which many research problems in robotics are framed through the same conceptual template: behavior is treated as the outward expression of hidden internal states, and successful interaction is framed as the ability to infer or reveal those states.

# The limits of ToM as a Foundation for Social Interaction in Robotics

A growing body of work challenges the idea of Theory of Mind as our primary means for understanding other people (Gallagher, 2001; Leudar & Costall, 2009; De Jaeger & Di Paolo, 2007). Instead, it is becoming increasingly clear that theory of mind explains only a subset of situations people find themselves in, and that we need to understand the problem of understanding other people not as an exercise of simulation or infer-ence but rather as a form of embodied practice (Gallagher, 2001). My critique here has a different aim. Rather than critiquing theory-of-mind research itself, my aim is to unpack the broader assumption that modeling hidden mental states should serve as the primary foundation for designing social interaction capabilities in robots.

Researchers in ethnomethodology and conversation analysis (EMCA) have long argued that meaning is not hidden behind behavior but rather a collaborative achievement (Garfinkel, 1967; Sacks, 1992). Ethnomethodology examines how meaning is produced in and through processes of social interaction. It examines the practical methods (ethnomethods) that participants of an interaction use to make their actions intelligible to one another. From this perspective, participants do not first infer what others intend and then respond. They respond to what the unfolding interaction suggests and act in ways that make their understanding publicly visible to others. Behavior does not come with preassigned meaning. For example, what counts as a greeting, an invitation or a misunderstanding is not determined by intent, nor the specific morphology of behavior in context, but rather emerges through the moment-to-moment sequence of actions and how they are responded to. Understanding, then, is not a process of inference, but an interactive practice. (Garfinkel, 1967; Sacks, 1992; Schegloff, 2007; Goodwin, 2018).

Research on participatory sense-making makes a similar claim. Rather than treating social sense-making as a process of modeling other minds, sense-making is seen as emergent from the coupling between agents (De Jaeger & Di Paolo, 2007). Participatory sense making has its roots in research on embodied cognition (Kirsh & Di Maglio, 1994) and enactivism (Varela, Rosch, & Thompson, 1991), which highlight the role of the body in cognition and view cognition as emergent from the interaction be-tween an organism and its environment. The approach reframes social cognition as emerging from the process of interaction between individuals in a social encounter, where the interaction process itself can take on a form of autonomy that individual are in engagement with. When people are engaging with each other they are working not to infer inner states but to keep the interaction going until it makes sense to all parties involved.

Recent developments in artificial intelligence underscore the idea that mental state inference is not required for social interaction. While turn-based, large language models produce responses that "make sense" without mental state inferences by merely predicting the most plausible next action. Recent work in robotics further demonstrates that effectively engaging in social interaction such as traffic for au-tonomous cars is achievable by modeling social dynamics rather than internal states (Sadigh et al., 2016) and the view that social cognition is limited to processes of the mind is becoming increasingly challenged in robotics (Prorok, 2025). These examples suggest that effective social coordination may arise from models that capture the dy-namics of interaction itself, pointing toward an alternative foundation for robotics grounded in coordination rather than mental-state inference.

Taken together, this work suggests that the assumptions of inside-out causality, mentalizing and sufficiency do not adequately capture the build of social engagements people find themselves in. Interaction is not best understood as a problem of inferring hidden states from observable behavior. It is better understood as a process of ongo-ing coordination where every move is a probe to shape and understand the unfolding of the interaction.

## An Interactional Foundation For Robotics

Focusing on theory of mind as a primary paradigm for social interaction risks us reaching a plateau for what robots can achieve and risks optimizing for situations in which agents must infer hidden states while underinvesting in the actual capacities required for the build of interactions that robots find themselves in. I therefore argue for an interactional foundation for social behavior in robotics—a paradigm grounded in ethnomethodology and conversation analysis as well as participatory sense-making that treats meaning as established through moment-to-moment coordination rather than inferred from hidden mental states. Adopting this perspective does not simply shift theoretical interpretation; it reframes core engineering questions in robotics, moving attention from architectures that reconstruct hidden states toward systems that learn policies for entering, sustaining, and repairing interaction.

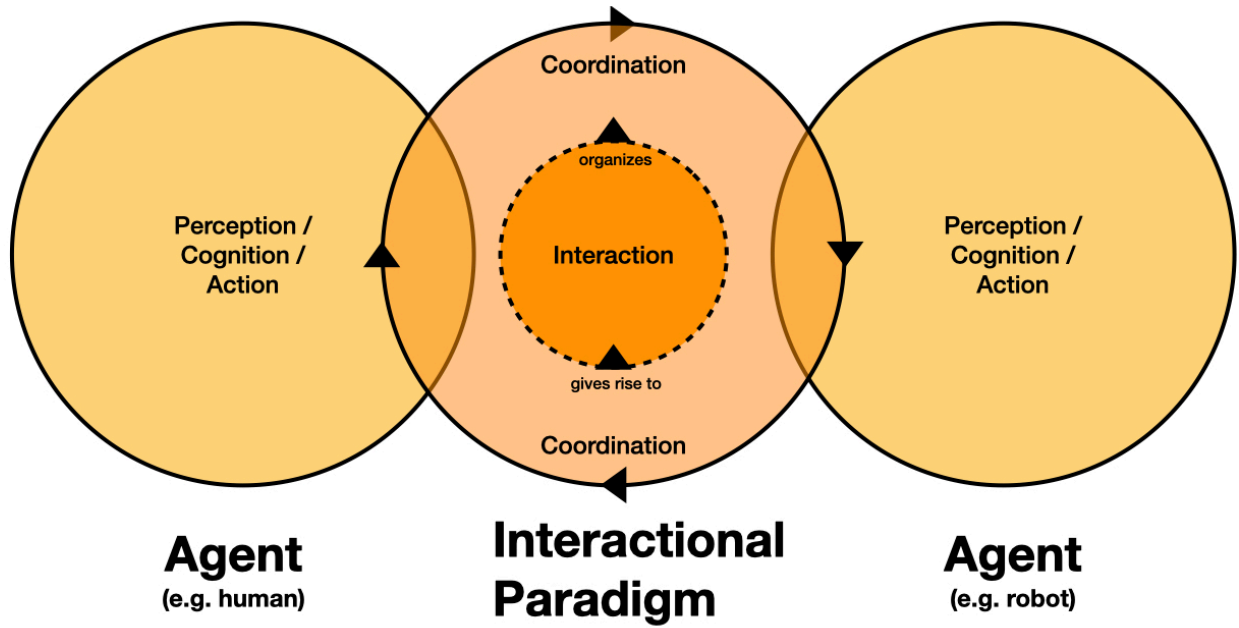

*Figure 2.* A conceptual visualization of two agents interacting with each other consistently with an interactional paradigm. Here, meaning making begins once the behavior of two agents becomes coupled as one agent starts responding to the other. The coordination of this coupling gives rise to an emergent interaction process that in turn organizes how coordination unfolds. Meaning is not decoded from behavior but produced between agents through ongoing mutual adjustment. Coordination and interaction are depicted as a circle equal in size to the agents to reflect the idea that interaction can take on a form of autonomy itself which shapes the trajectory of an interaction. Dashed boundaries indicate that agents and the interaction process are open and mutually constitutive rather than self-contained. *Note.* I am using a dyadic interaction example here only for illustration. The same paradigm holds for multi-party interactions.

**From inside-out to outside-in:** The inside-out logic assumes that the best way for a robot to generate an appropriate response is to develop models of internal states and processes that can reason about an input and produce a corresponding output. Examples of this are cognitive architectures or emotion generation models grounded in basic emotion theory.

An interactional foundation, however, holds that social behavior should be understood as response to situational demands rather than internal states. Actions are not chosen because they express inner states, but because the situation makes them relevant. For example, a "thank you" in response to a nice gesture should be less understood as a behavior caused by the internal state of gratitude and better be understood as the most plausible response to a social situation that made such a response particularly relevant. The field thus needs a shift from building internal architectures (e.g. of cognition and emotion) towards models that predict plausible next actions. While not truly participatory, a powerful example of this approach are recent achievements in replacing mental-state inference with techniques that rely on large language models to suggest plausible next actions for robots (Mahadevan et al., 2024).

**From mentalizing to participation:** To understand social behavior, current approaches predominantly rely on second hand approaches for sense making: take a discrete stream of data such as a video or sensor data, decode or infer its meaning, and then generate an appropriate response. Reversely, to make robot behavior understandable by people it is assumed that the behavior a robot exhibits needs to be inherently legible or explainable. An interactional foundation, however, implies that for a robot to under-stand a social interaction it needs to participate in it and reversely that explainability and legibility have to be sought in interaction rather than in the morphologies of discrete behaviors.

Such a shift implies a re-focusing from capabilities for scene understanding and internal state inference towards focusing on capabilities that enable robots to engage, participate and sustain interaction over time. An interaction perspective further understands behavior and sense-making as continuous, not as sequences of discrete steps of observation, meaning making, and responding. It requires a move from turn based models of social interaction towards models of continuous correspondence. This shift understands interaction as means to sustain and enable sense-making rather than an end in and of itself. Building interaction capabilities for robots requires approaches that allow for the emergence of understanding over time rather than ad hoc from a given input - capabilities that enable robots to enter and sustain a process of correspondence with another actor and models that continuously update what the interaction turns out to be about. Computationally, this suggests shifting from architectures that infer hidden states toward systems that learn policies for sustaining coordination—e.g., policies optimized for responsiveness, contingency detection, and repair when interactional expectations are violated.

Recent work has used generative AI models to parse behavior in order to generate plausible responses (e.g. Mahadevan et al., 2024). While such approaches demonstrate capabilities for robots to generate appropriate robots without internal state inferences, these approaches still assume that meaning can be derived from the perspective of an un-involved observer. Alternatively, work by Sadigh and colleagues (2016) has shown how we can build interaction models for autonomous cars that build social understanding through participation in interaction rather than through the mere observation of it. This work is promising as it understands sense-making as a participatory process, but such approaches are currently limited to specific tasks such as a lane change for an autonomous car or a robot handover rather than formulating general frameworks for interaction and participatory sense making.

**From sufficiency to meaning potential:** Dominant approaches for behavior generation and interpretation assume that there is stability in the meaning of behavior. It is assumed that we can imbue robot behaviors with inherent legibility, expressiveness or explainability and that we can classify behavior into fixed and stable categories of meaning such as a greeting, a handover, a smile or takeover attempt. Understanding behavior thus becomes a data availability problem and generating behavior becomes problem of generating behaviors with inherent legibility.

An interactional foundation, however, suggests that the meaning of behavior is not inherent in the morphology of the behavior no matter how much contextual information or "noise" we might take into account. From an interactional perspective, any moment in a stream of behavior can carry social meaning if it is responded to as such. That means that certain behaviors are likely to be understood in a specific way but we can never be sure until we see how they are taken up. Even canonical gestures such as a hand-wave or a nod should be understood as having meaning potential rather than fixed meaning.

This has direct consequences for how we should think about robot action. Current approaches to behavior generation treat a robot's response or action as the end point of a reasoning process. For example, a robot recognizes a an opening in traffic, reasons it is large enough to merge and then plans an appropriate path. Or a robot recognizes an intent to greet and executes a greeting action in response. In both cases, the response is treated as the final output of a perception and inference process. From an interactional perspective, however, developing capabilities for robots to participate in interaction requires a shift from treating a robot's actions as purely pragmatic towards treating them as both pragmatic and epistemic.

Kirsh and Maglio (1994) famously distinguished pragmatic actions - aimed at changing the world towards a desired goal - from epistemic actions - aimed at revealing information. Applied to robotics, this means that behavior should never be seen just as an output but also as a probe to gather information. In other words we should understand behavior as a hypothesis. Everything a robot does moves the interaction forward but also proposes a potential interpretation of the situation. The robot must wait for how its behavior is taken up and be able to respond if it is not responded to as intended. This shifts behavior design away from treating actions as final products and towards treating them as interactional moves whose meaning is only stabilized through response.

## Conclusion

Rather than asking how robots can infer what others think or intend, an interactional perspective suggests a different central challenge for robotics: designing systems capable of entering into processes of coordination through which social meaning can emerge.